\documentclass[lettersize,journal]{IEEEtran}

\usepackage{amsmath,amsfonts}
\usepackage{algorithmic}
\usepackage{algorithm}
\usepackage{array}
\usepackage[caption=false,font=normalsize,labelfont=sf,textfont=sf]{subfig}
\usepackage{textcomp}
\usepackage{stfloats}
\usepackage{url}
\usepackage{verbatim}
\usepackage{graphicx}
\usepackage[style=ieee]{biblatex}
\bibliography{references}
\usepackage{orcidlink}

\everydisplay{\footnotesize}
\everymath{\footnotesize}
\usepackage{xcolor}
\usepackage{pifont}
\usepackage{booktabs}
\usepackage{multirow}
\usepackage{hyperref}
\hypersetup{hidelinks}
\usepackage{newtxmath}

\definecolor{darkgreen}{rgb}{0.1, 0.6, 0.1}
\definecolor{darkred}{rgb}{0.8, 0.2, 0.2}
\newcommand{\cmark}{\textcolor{darkgreen}{\ding{51}}}
\newcommand{\xmark}{\textcolor{darkred}{\ding{55}}}

\begin{document}

\title{\textsc{InsightX Agent}: An LMM-based Agentic Framework with Integrated Tools for Reliable X-ray NDT Analysis}

\author{Jiale Liu$^{\orcidlink{0009-0007-5881-3118}}$, Huan Wang$^{\orcidlink{0000-0002-1403-5314}}$*, Yue Zhang, Xiaoyu Luo$^{\orcidlink{0009-0002-1600-9512}}$, Jiaxiang Hu$^{\orcidlink{0009-0002-7874-4336}}$, Zhiliang Liu$^{\orcidlink{0000-0002-4133-8230}}$*,~\IEEEmembership{Senior Member,~IEEE} \\ and Min Xie$^{\orcidlink{0000-0002-8500-8364}}$,~\IEEEmembership{Fellow,~IEEE}
\thanks{This work was supported in part by the Chengdu Science and Technology Program (Grant No. 2025-YF12-00016-RC), in part by the Research Grant Council of Hong Kong (Grant Nos. CityU JRFS2526-1S09, 11201023, 11202224), and in part by the National Natural Science Foundation of China (Grant No. 72371215). (Corresponding authors: Huan Wang, e-mail: hwan36@cityu.edu.hk, and Zhiliang Liu, e-mail: zhiliang\_liu@uestc.edu.cn)}%
\thanks{Jiale Liu is with the School of Physics and Astronomy, The University of Edinburgh, Edinburgh, EH9 3FD, United Kingdom (e-mail: Jiale.Liu@ed.ac.uk).}%
\thanks{Yue Zhang, Xiaoyu Luo, and Jiaxiang Hu are with the Glasgow College, University of Electronic Science and Technology of China, Chengdu, Sichuan, 611731, China.}
\thanks{Zhiliang Liu is with the School of Mechanical and Electrical Engineering, University of Electronic Science and Technology of China, Chengdu, Sichuan, 611731, China.}
\thanks{Huan Wang and Min Xie are with the Department of Systems Engineering, City University of Hong Kong, Hong Kong, China}%
\thanks{This work has been submitted to the IEEE for possible publication. Copyright may be transferred without notice, after which this version may no longer be accessible.}
}

\markboth{Accepted by IEEE Transactions on Reliability}%
{Liu \MakeLowercase{\textit{et al.}}: InsightX Agent: An LMM-based Agentic Framework for Reliable X-ray NDT Analysis}

\maketitle

\begin{abstract}
Non-destructive testing (NDT), particularly X-ray inspection, is vital for industrial quality assurance, yet existing deep-learning-based approaches often lack interactivity, interpretability, and the capacity for critical self-assessment, limiting their reliability and operator trust. To address these shortcomings, this paper proposes \textsc{InsightX Agent}, a novel LMM-based agentic framework designed to deliver reliable, interpretable, and interactive X-ray NDT analysis. Unlike typical sequential pipelines, \textsc{InsightX Agent} positions a Large Multimodal Model (LMM) as a central orchestrator, coordinating between the Sparse Deformable Multi-Scale Detector (SDMSD) and the Evidence-Grounded Reflection (EGR) tool. The SDMSD generates dense defect region proposals from multi-scale feature maps and sparsifies them through Non-Maximum Suppression (NMS), optimizing detection of small, dense targets in X-ray images while maintaining computational efficiency. The EGR tool guides the LMM agent through a chain-of-thought-inspired review process, incorporating context assessment, individual defect analysis, false positive elimination, confidence recalibration and quality assurance to validate and refine the SDMSD's initial proposals. By strategically employing and intelligently using tools, \textsc{InsightX Agent} moves beyond passive data processing to active reasoning, enhancing diagnostic reliability and providing interpretations that integrate diverse information sources. Experimental evaluations on the GDXray+ dataset demonstrate that \textsc{InsightX Agent} not only achieves a high object detection F1-score of 96.54\% but also offers significantly improved interpretability and trustworthiness in its analyses, highlighting the transformative potential of LMM-based agentic frameworks for industrial inspection tasks.
\end{abstract}

\begin{IEEEkeywords}
Large Multimodal Model, Agentic Framework, Non-Destructive Testing, X-ray Inspection.
\end{IEEEkeywords}

\section{Introduction}
\IEEEPARstart{N}{on-destructive} testing (NDT) plays an indispensable role in ensuring the safety, reliability, and operational integrity of critical components across a multitude of industrial sectors, including aerospace, manufacturing, and construction \cite{ines_silva_review_2023, li_reliability_2021, harmouche_statistical_2016}. By enabling the evaluation of material and structural integrity without causing damage, NDT methodologies are fundamental to quality assurance, preventative maintenance, and risk mitigation \cite{Ou2021Recent, li_reliability_2021}. Among the diverse NDT techniques, X-ray radiography stands out for its unique capability to reveal internal defects that are otherwise invisible to surface inspection methods, including voids, cracks, and inclusions \cite{czyzewski_detecting_2021, wu_ameliorated_2021}. This penetrative power makes X-ray testing particularly important for assessing the internal quality of castings, welds, and complex assemblies, where internal flaws can lead to catastrophic failures \cite{du_approaches_2019}. 

Typically, the interpretation of X-ray images is often performed by trained inspectors who interpret the images based on their experience and knowledge of the materials and structures being examined \cite{mery_gdxray_2015, mery_aluminum_2021}. However, the manual interpretation of X-ray radiographs is inherently labor-intensive, subjective, and heavily reliant on the expertise and vigilance of human inspectors. These factors can lead to inconsistencies in defect identification and characterization, posing limitations on throughput and overall inspection reliability.

In recent years, with the advent of data-driven approaches, deep learning (DL) has emerged as a powerful tool for automating the analysis of X-ray images and other related industrial signals \cite{Vandewalle2024Machine, gu_lightweight_2024}. These models can learn complex patterns and features from large datasets, enabling them to identify defects with high accuracy and speed. For instance, convolutional neural networks (CNNs) have been extensively investigated in this domain. Xu \textit{et~al.} \cite{Xu2024High-performance} developed a modified U-Net-based segmentation model for X-ray computed tomography, achieving 84.83\% Mean Intersection over Union and 98.87\% pixel accuracy for pore and crack detection in aluminum alloy samples. Additionally, Shen \textit{et~al.} \cite{Shen2021Automatic} introduced a two-stream CNN architecture that processes both original and CLAHE-enhanced X-ray images with a gated multilayer fusion module, achieving 42.2 mIoU and 59.2 Dice scores for defect segmentation tasks. Zhang \textit{et~al.} \cite{Zhang2020Automated} enhanced Feature Pyramid Network (FPN) architecture with DetNet and Path Aggregation Network to address feature imbalance and location information loss, incorporating soft Non-Maximum Suppression (NMS) and soft Intersection Over Union evaluation criteria for improved casting defect detection performance. 

Beyond conventional CNN architectures, researchers have also explored more advanced techniques to enhance X-ray defect detection capabilities. Wang \textit{et~al.} \cite{wang_self-attention_2020} introduced self-attention mechanisms within their detection framework, demonstrating superior performance in identifying subtle defects within aluminum alloy castings against complex backgrounds. To address the challenge of limited labeled data in industrial settings, Intxausti \textit{et~al.} \cite{intxausti_arbaiza_methodology_2024} demonstrated that self-supervised pretraining methods using X-ray images achieved superior results compared to standard ImageNet-pretrained models. Similarly, transfer learning has emerged as another effective strategy for addressing data scarcity challenges \cite{chen_transfer_2024}. Wen \textit{et~al.} \cite{Wen2025Data-Efficient} proposed a domain adaptation framework for X-ray PCB defect detection, introducing the XD-PCB dataset and achieving a 10\% improvement in average precision compared to other adaptation methods. Zhang \textit{et~al.} \cite{zhang_intelligent_2024} developed a transfer learning network incorporating convolutional autoencoders for lithium-ion battery structural health monitoring using X-ray CT images, establishing a structural deviation index that significantly improved the generalization ability of their detection model. 

Despite their detection prowess, most existing AI-based NDT systems function as pure ``black box'' systems. They output bounding box coordinates and confidence scores that necessitate further post-processing to derive actionable insights, but crucially, they often provide limited explanation as to why a specific region was flagged as defective \cite{Taylor2020Defect}. This lack of transparency and interpretability can hinder operator trust and makes it difficult to integrate these systems seamlessly into complex decision-making workflows \cite{liu2024braininspired, tian_toward_2024, wang_auto-embedding_2024}. Furthermore, current systems typically lack interactivity. Operators cannot easily query the system for further details, ask for clarification on ambiguous detections, or engage in a dialogue to understand the potential implications of identified flaws. In fact, industrial NDT requires systems that not only detect defects with high precision but also offer contextual, interpretable insights and facilitate a deeper understanding of the findings, especially for operators with varying levels of NDT expertise.

Recent research has begun to realize this potential through pioneering applications of LMMs in anomaly detection \cite{liu_aerogpt_2025}. Gu \textit{et~al.} \cite{gu_anomalygpt_2024} developed AnomalyGPT, an innovative approach that fine-tuned Large Vision-Language Models using simulated anomalous images and corresponding textual descriptions, achieving state-of-the-art performance with 86.1\% accuracy and 94.1\% image-level AUC on the MVTec-AD dataset while eliminating manual threshold adjustments. Cao \textit{et~al.} \cite{cao_towards_2023} explored GPT-4V's generic anomaly detection capabilities across multiple modalities, including images, videos, point clouds, and time series data, demonstrating its effectiveness in zero-shot detection through enhanced prompting strategies that incorporated class information and human expertise. Building upon this foundation, Zhang \textit{et~al.} \cite{zhang_gpt-4v-ad_2024} specifically investigated GPT-4V's potential for visual anomaly detection through a Visual Question Answering paradigm, proposing a comprehensive framework with granular region division and prompt design components that achieved 77.1\% image-level AU-ROC on the MVTec AD dataset. These pioneering studies establish the foundational framework for integrating conversational AI capabilities into industrial inspection systems.

However, merely leveraging LMMs as standalone solutions for industrial X-ray NDT reveals inherent limitations. Firstly, while LMMs can be trained for visual grounding tasks like outputting defect coordinates, achieving the requisite precision for industrial applications demands extensive, pixel-perfect annotated datasets \cite{zhang_llava-grounding_2023, ye_moat_2025}. Such datasets are difficult and costly to acquire in specialized industrial scenarios, rendering this approach often impractical. Furthermore, direct visual grounding may suffer from hallucinations, which are the generation of plausible-sounding but factually incorrect or unsubstantiated defect localizations \cite{huang_survey_2025, favero_multi-modal_2024}. Secondly, even with domain-specific fine-tuning, using an LMM as a passive, end-stage analyzer of information (such as raw images or preprocessed detector outputs) means it might still inherit upstream inaccuracies or fail to critically engage with the primary visual evidence. The true transformative potential of LMMs in these critical applications extends beyond their direct identification or analytical power; it lies in their capacity to orchestrate a more holistic, critically self-aware, and tool-augmented diagnostic process.

To address these limitations, this paper proposes \textsc{InsightX Agent}, a novel LMM-based agentic framework for reliable, interpretable, and interactive X-ray NDT analysis. {Typically, an agentic system is characterized by its ability to autonomously reason, plan, and utilize tools to achieve complex goals, moving beyond simple input-output processing. Thus, unlike typical sequential pipelines, \textsc{InsightX Agent} positions a Large Multimodal Model (LMM) as a central orchestrator coordinating between two key tools: the Sparse Deformable Multi-Scale Detector (SDMSD) and the Evidence-Grounded Reflection (EGR) tool. The SDMSD, drawing inspiration from recent advancements in transformer-based object detectors, generates dense defect region proposals across multi-scale feature maps, and then sparsifies them through NMS. This dense-to-sparse strategy, combined with the computational efficiency of deformable attention mechanisms \cite{zhu_deformable_2020}, is specifically designed to optimize the detection of the small, densely clustered targets characteristic of X-ray images.} {The EGR tool is a novel mechanism we propose to harness the wide range of knowledge and advanced reasoning capabilities latent within LMMs. Inspired by the emergent cognitive abilities demonstrated in recent LMM research, such as multi-step reasoning via Chain-of-Thought \cite{wei_chain--thought_2022} and iterative self-refinement \cite{madaan_self-refine_2023}, we designed the EGR protocol to structure this capability for the high-stakes domain of NDT. It guides the agent through a purpose-built sequence of context assessment, individual defect analysis, false positive elimination, confidence recalibration, and quality assurance, enabling it to systematically validate and refine the SDMSD's initial proposals. This strategy leverages LMM's cognitive capabilities without requiring pixel-perfect localization training. By strategically employing the tools, \textsc{InsightX Agent} moves beyond passive data processing to active reasoning, enhancing diagnostic reliability and providing interpretations that integrate diverse information sources. Experimental evaluations on the GDXray+ dataset demonstrate that \textsc{InsightX Agent} achieves a high object detection F1-score of 96.54\% while offering significantly improved interpretability and trustworthiness, highlighting the transformative potential of LMM-based agentic frameworks for industrial inspection tasks.}

\begin{table*}[htbp]
\centering
\caption{Capability Comparison of Different NDT Analysis Approaches}
\label{tab:paradigm_comparison}
\begin{tabular}{@{}lccccc@{}}
\toprule 
\textbf{Comparison Dimension} & \begin{tabular}[c]{@{}c@{}}Traditional CV \\ Methods\end{tabular} & \begin{tabular}[c]{@{}c@{}}Standard DL \\ Detectors\end{tabular} & \begin{tabular}[c]{@{}c@{}}LMM Direct \\ Grounding\end{tabular} & \begin{tabular}[c]{@{}c@{}}LMM-based \\ Workflows\end{tabular} & \textbf{\begin{tabular}[c]{@{}c@{}}InsightX Agent \\ (Ours)\end{tabular}} \\
\midrule
High-Precision Defect Detection & \xmark & \cmark & \xmark & \cmark & \textbf{\cmark} \\
Reflection-based Result Optimization & \xmark & \xmark & \xmark & \xmark & \textbf{\cmark} \\
Result Interpretability & Medium & Low & High & High & \textbf{High} \\
User Intent Recognition & \xmark & \xmark & \xmark & \xmark & \textbf{\cmark} \\
Versatile Task Handling & \xmark & \xmark & \cmark & \xmark & \textbf{\cmark} \\
Interactive Dialogue & \xmark & \xmark & \cmark & \cmark & \textbf{\cmark} \\
Required Expertise Level & High & Medium & Low & Low & \textbf{Low} \\
\bottomrule
\end{tabular}
\end{table*}

As shown in \autoref{tab:paradigm_comparison}, our proposed \textsc{InsightX Agent} framework offers significant advantages over existing NDT analysis approaches across multiple dimensions. While traditional computer vision methods and standard deep learning detectors excel in certain areas, they lack the interpretability and interactive capabilities that are crucial for modern industrial applications. LMM-based approaches improve interpretability but often compromise on detection precision or versatility. In contrast, \textsc{InsightX Agent} uniquely combines high-precision defect detection with reflection-based result optimization, comprehensive user intent recognition, and versatile task handling capabilities.

The main contributions of this paper can be summarized as follows:
\begin{enumerate}
    \item {We propose \textsc{InsightX Agent}, a novel LMM-based agentic framework for NDT that establishes a new paradigm of active, tool-driven reasoning over passive data processing. The framework utilizes an LMM-based agent that is domain-adapted for NDT and orchestrates specialized tools for different purposes of analysis.}
    \item {We design and implement two key components for the agentic framework: (1) the Sparse Deformable Multi-Scale Detector (SDMSD) for efficient and precise defect localization in X-ray imagery, and (2) the Evidence-Grounded Reflection (EGR) mechanism, a structured protocol that enables the LMM agent to critically validate and refine initial detections against visual evidence.}
    \item {Through both quantitative and qualitative experiments on the GDXray+ dataset, we demonstrate that \textsc{InsightX Agent} not only achieves state-of-the-art detection accuracy but also provides significantly enhanced interpretability and reliability, showcasing the transformative potential of our agentic approach for industrial inspection workflows.}
\end{enumerate}

{The remainder of this paper is structured as follows: Section II introduces the methodology of the proposed \textsc{InsightX Agent} framework, including the overview of the agentic framework architecture, domain adaptation strategies for the LMM agent core, SDMSD for efficient defect localization, and EGR mechanism for diagnostic validation, and training optimization procedures; Section III presents experimental validation on the GDXray+ dataset, including ablation studies, comparative analysis with existing approaches and studies to show the agentic and contextual capabilities of LMMs, as well as several discussions; Section IV concludes the paper by summarizing the key contributions and findings.}

\section{Methodology}
\begin{figure*}[htbp] 
    \centering
    \includegraphics[width=\textwidth]{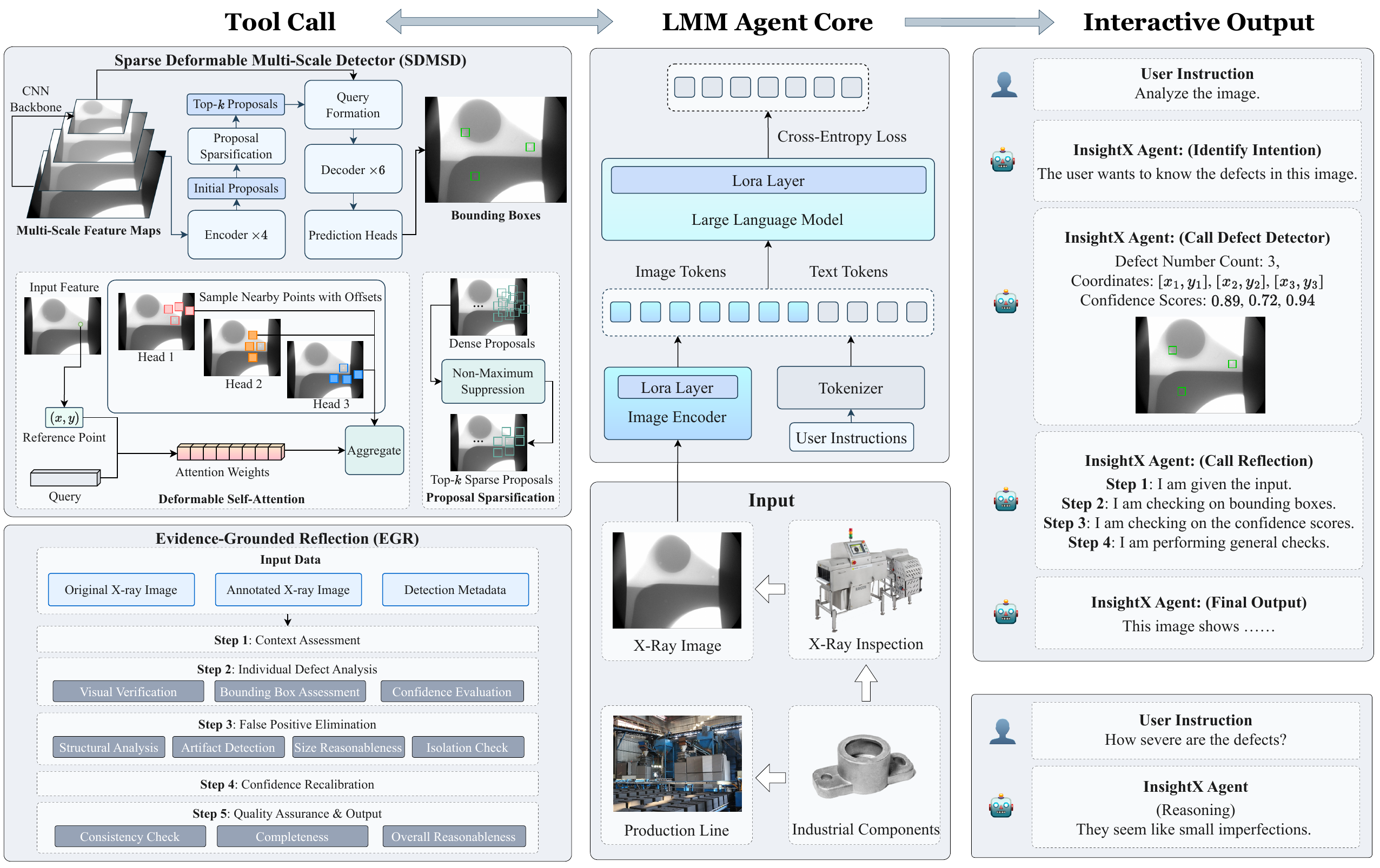} 
    \caption{The architecture of \textsc{InsightX Agent}. Industrial components undergo X-ray inspection, generating images that serve as input. The LMM agent core orchestrates the process and calls necessary tools to perform tasks instructed by users.}
    \label{fig:insightx_agent_framework}
\end{figure*}

\subsection{Overview of Agentic Framework}
{The \textsc{InsightX Agent} framework, as illustrated in \autoref{fig:insightx_agent_framework}, presents an LMM-based agentic system that positions the LMM as a central orchestrator rather than a passive end-stage interpreter. Given an X-ray image and an optional user query, the LMM agent core performs implicit intention recognition to determine the appropriate analytical pathway. For defect identification tasks, the system invokes two specialized tools: the SDMSD for precise defect localization, and the EGR mechanism for systematic validation.}

{The SDMSD integrates transformer-based processing with proposal sparsification and deformable attention mechanisms. It generates dense object proposals from multi-scale feature maps, applies NMS filtering to establish a refined sparse candidate set, and processes them through deformable self-attention to achieve efficient localization. Critically, these detector proposals are treated as hypotheses requiring validation rather than definitive diagnoses.}

{The LMM agent core then invokes the EGR mechanism to critically evaluate detector proposals against the original radiographic imagery and embedded NDT knowledge. This structured reasoning protocol encompasses context assessment, individual defect analysis, false positive elimination, confidence recalibration, and quality assurance, enabling autonomous validation, refinement, or uncertainty flagging. For general queries not requiring defect analysis, the LMM agent core leverages domain knowledge to generate direct responses without tool invocation. This design showcases the high reliability, interpretability, and interactivity required in modern industrial NDT applications.}

\subsection{Domain Adaptation for Agentic Core}

While LMMs have demonstrated exceptional performance across general-purpose domains, they exhibit limited specialized knowledge in industrial applications, particularly within NDT and defect analysis. This limitation leads to reduced effectiveness for NDT-specific tasks, including accurate defect identification, precise localization, and reliable assessment of structural anomalies in radiographic imagery. The technical terminology, visual patterns, and diagnostic reasoning protocols required for industrial inspection differ substantially from scenarios encountered during pre-training, necessitating targeted domain adaptation.

To enable effective utilization in industrial inspection, Low-Rank Adaptation (LoRA) is employed as the primary domain adaptation strategy \cite{hu_lora_2021}. {This approach is chosen over full fine-tuning, which is computationally prohibitive for 7B-parameter models (requiring well over 80GB of VRAM for standard training with gradients and optimizer states) and risks catastrophic forgetting of the model's general reasoning abilities.} LoRA operates on the principle that task-specific adaptations can be effectively captured through low-rank matrix decompositions. Given a pre-trained weight matrix $W_0 \in \mathbb{R}^{d \times k}$, LoRA introduces trainable rank decomposition matrices such that the adapted weights become:
\begin{equation}
W = W_0 + \Delta W = W_0 + BA
\end{equation}
where $B \in \mathbb{R}^{d \times r}$, $A \in \mathbb{R}^{r \times k}$, and $r \ll \min(d, k)$ represents the intrinsic rank. This decomposition reduces the number of trainable parameters from $d \times k$ to $r \times (d + k)$, enabling efficient domain adaptation while preserving pre-trained knowledge.

The domain adaptation strategy encompasses a two-stage curriculum designed to instill both declarative NDT knowledge and procedural reasoning capabilities. The first stage involves constructing a comprehensive NDT knowledge corpus comprising systematically organized question-answer pairs covering fundamental concepts, terminology, and diagnostic principles. This corpus is formally represented as:
\begin{equation}
\mathcal{D}_{\text{knowledge}} = \bigcup_{i=1}^{N} \{(q_i, a_i, m_i)\}
\end{equation}
where $q_i$ represents domain-specific queries, $a_i$ denotes human-validated responses, and $m_i$ encodes categorical metadata including defect taxonomies, material properties, and inspection methodologies. {Specifically, we constructed a knowledge corpus containing 1,000 question-answer pairs covering common NDT issues and operational guidelines. For instance, typical queries include ``What morphological characteristics does a porosity defect typically exhibit in X-ray images of aluminum castings?'' with corresponding human-validated responses such as ``Porosity typically manifests as circular or elliptical dark regions with clear boundaries, indicating areas of reduced material density caused by gas inclusions.'' Through fine-tuning on this corpus, the LMM acquires core NDT terminology, fundamental principles, and diagnostic knowledge essential for industrial inspection tasks.}

The second stage employs a template-driven alignment mechanism designed to align model output patterns with industrially relevant, defect-centric analytical frameworks. Rather than relying on complex preference optimization approaches requiring extensive human annotation, an exemplar-guided learning paradigm is utilized that leverages carefully crafted response templates to establish consistent analytical patterns. This alignment process is formulated as a structured imitation learning problem:
\begin{equation}
\mathcal{L}_{\text{template}} = \mathbb{E}_{(\mathbf{x},y_{\text{template}}) \sim \mathcal{D}_{\text{exemplar}}} \left[ -\log P_\theta(y_{\text{template}}|\mathbf{x}) \right]
\end{equation}
where $y_{\text{template}}$ represents human-crafted responses following the structured analytical framework, and $\mathcal{D}_{\text{exemplar}}$ denotes the dataset of template-response pairs. {This template-driven alignment is fundamental to ensuring that the agent's output is structured, reliable, and, critically, quantitatively evaluable for industrial deployment. The template-driven alignment dataset was constructed through systematic curation of approximately 100 representative X-ray images from the GDXray+ training set. For each selected image, comprehensive analytical reports were generated through cross-evaluation using commercial large language models (Gemini 2.5 Pro and GPT-4o), followed by human integration and verification. These exemplar reports demonstrate the structured analytical format, evidence-grounded reasoning patterns, and industry-standard diagnostic protocols required for reliable NDT analysis. Through imitation learning on these high-quality exemplars, the LMM learns not only domain knowledge but, more importantly, how to organize and present analytical findings in a structured, rigorous, and industry-compliant manner.}

The response template follows three core principles to ensure reliable NDT diagnostic outputs from the LMM agent. First, evidence-based grounding requires all diagnostic statements to reference specific visual features observed in the X-ray imagery, preventing unsupported conclusions. Second, quantitative precision mandates measurable defect characteristics (size, location, severity) rather than vague qualitative descriptions, enabling consistent diagnostic reproducibility. Third, actionable recommendations ensure all suggested next steps align with common industry inspection standards. Each template response follows this structured analytical framework:
\begin{equation}
\mathcal{T}_{\text{response}} = \langle \text{Detection}, \text{Evidence}, \text{Assessment}, \text{Recommendation} \rangle
\end{equation}

To reinforce consistent behavioral patterns during training, a response regularization strategy is employed that penalizes deviations from established template structures while maintaining semantic flexibility:
\begin{equation}
\mathcal{L}_{\text{reg}} = \alpha \cdot \mathcal{D}_{\text{KL}}(P_\theta(y|\mathbf{x}) \parallel P_{\text{template}}(y|\mathbf{x}))
\end{equation}
where $\mathcal{D}_{\text{KL}}(P \parallel Q)$ represents the Kullback-Leibler divergence measuring the distributional difference between model outputs and template patterns, $\alpha$ controls the strength of structural adherence, and $P_{\text{template}}$ represents the learned distribution over valid template patterns.

The integration of LoRA adaptation with the structured training curriculum yields an LMM agent core that combines the broad reasoning capabilities of foundation models with specialized knowledge and disciplined analytical patterns essential for reliable NDT analysis. This approach enhances domain expertise while establishing a robust cognitive foundation for sophisticated tool orchestration and evidence-based reasoning mechanisms that characterize \textsc{InsightX Agent}.

\subsection{Sparse Deformable Multi-Scale Detector}

To facilitate more accurate and precise anomaly detection, a separate defect detector, namely the Sparse Deformable Multi-Scale Detector (SDMSD), is integrated into the framework.

The SDMSD architecture initiates with comprehensive dense proposal generation from multi-scale CNN feature maps, ensuring exhaustive coverage of potential anomalies regardless of scale or subtlety. This initial stage prioritizes maximum recall to capture the complete spectrum of defect manifestations characteristic of industrial radiographic imagery. Subsequently, the detector implements a sparsification approach utilizing NMS to eliminate redundant candidates and establish a refined proposal set. This process can be mathematically formulated as:
\begin{equation}
\begin{aligned}
\mathcal{S}_{\text{sparse}} &= \text{NMS}(\mathcal{B}_{\text{dense}}, \tau_{\text{IoU}}) \\
&= \{\mathbf{b}_i \in \mathcal{B}_{\text{dense}} : \forall \mathbf{b}_j \in \mathcal{B}_{\text{dense}}, \\
&\quad c_j > c_i \Rightarrow \text{IoU}(\mathbf{b}_i, \mathbf{b}_j) < \tau_{\text{IoU}}\}
\end{aligned}
\end{equation}
where $\mathcal{B}_{\text{dense}}$ represents the initial dense proposal set, $c_i$ denotes the confidence score of proposal $\mathbf{b}_i$, $\tau_{\text{IoU}}$ is the intersection-over-union threshold, and $\mathcal{S}_{\text{sparse}}$ is the resulting sparse proposal set. This NMS-driven refinement transforms the dense candidate field into a sparse, high-quality collection that concentrates analytical resources on the most promising defect hypotheses—a critical advantage for resolving small and densely clustered anomalies prevalent in X-ray inspection scenarios.

Following the sparsification process, the refined sparse proposal ensemble $\mathcal{S}_{\text{sparse}}$ serves as input to a transformer encoder-decoder architecture employing multi-scale deformable attention. This two-stage approach ensures that computational resources are concentrated on the most promising defect candidates while maintaining comprehensive spatial coverage. For each query element $q$ with content feature $z_q$ and reference point $p_q$, the deformable attention mechanism selectively samples features from $K$ learnable sampling points:
\begin{equation}
\text{DeformAttn}(z_q, p_q, \mathbf{X}) = \sum_{m=1}^{M} \mathbf{W}_m \left[ \sum_{k=1}^{K} \mathbf{W}_{mqk} \cdot \mathbf{W}'_m \mathbf{X}(p_q + \Delta p_{mqk}) \right]
\end{equation}
where $M$ denotes attention heads, $\Delta p_{mqk}$ represents learnable 2D offsets, and $\mathbf{W}_{mqk}$ are attention weights. This selective sampling strategy achieves superior computational efficiency by focusing exclusively on informative spatial locations identified through the sparse proposal generation process.

To enable comprehensive information aggregation across feature hierarchies, the multi-scale extension processes the sparse queries at multiple feature levels:
\begin{equation}
\begin{aligned}
&\text{MSDeformAttn}(z_q, \hat{p}_q, \{\mathbf{X}^l\}_{l=1}^{L}) = \\
&\quad \sum_{m=1}^{M} \mathbf{W}_m \left[ \sum_{l=1}^{L} \sum_{k=1}^{K} \mathbf{W}_{mlqk} \cdot \mathbf{W}'_m \mathbf{X}^l(\phi_l(\hat{p}_q) + \Delta p_{mlqk}) \right]
\end{aligned}
\end{equation}
where $\hat{p}_q$ denotes normalized coordinates and $\phi_l(\cdot)$ maps them to the $l$-th feature level.

The decoder iteratively refines bounding box predictions and classification confidence through cross-attention mechanisms with encoder features. Training optimization employs set-based Hungarian matching algorithms to establish an optimal assignment between sparse proposals and ground truth annotations, enabling end-to-end learning without manual correspondence specification.

The SDMSD demonstrates significant advantages over conventional detection architectures by combining more comprehensive coverage through both dense and sparse query generation, and computational efficiency through deformable attention.

\subsection{Evidence-Grounded Reflection}

To address false positive mitigation and diagnostic confidence calibration in automated X-ray defect detection, we propose the Evidence-Grounded Reflection (EGR) mechanism, which provides a systematic validation framework that implements cognitively-inspired verification protocols with enhanced bounding box refinement capabilities, structured rejection criteria, and comprehensive quality control mechanisms to ensure rigorous evidential scrutiny of detection outputs prior to diagnostic commitment.

The EGR mechanism operates through a six-stage sequential validation process that systematically evaluates detector proposals against visual evidence, domain knowledge, and established NDT principles. For a given set of detection proposals $\mathcal{D} = \{d_1, d_2, ..., d_n\}$ where each detection $d_i$ comprises spatial coordinates $\mathbf{b}_i = (x, y, w, h)$ and confidence score $c_i$, the framework executes the following structured analytical cascade:
\begin{equation}
\begin{aligned}
\text{EGR}(\mathcal{D}, I, I^*) &= \text{Context}(I, I^*) \\
&\rightarrow \text{Analyze}(\mathcal{D}, I, I^*) \\
&\rightarrow \text{Filter}(\mathcal{D}) \\
&\rightarrow \text{Recalib}(\mathcal{D}) \\
&\rightarrow \text{QualityCheck}(\mathcal{D}) \\
&\rightarrow \text{Generate}(\mathcal{D})
\end{aligned}
\end{equation}
where $I$ represents the original radiographic image and $I^*$ denotes the annotated image with bounding box overlays.

The context assessment stage establishes the foundational understanding by evaluating image quality characteristics and identifying systematic patterns that may affect subsequent analysis. The individual detection analysis stage performs comprehensive evaluation for each proposal through three sub-components: visual verification, bounding box assessment, and confidence evaluation. The visual verification function is formalized as:
\begin{equation}
\mathcal{V}(d_i, I, I^*) = \langle \mathcal{A}(\mathbf{b}_i, I), \mathcal{Q}(\mathbf{b}_i, I^*), \mathcal{C}(\mathbf{b}_i, I) \rangle
\end{equation}
where $\mathcal{A}(\mathbf{b}_i, I)$ evaluates anomaly presence and consistency, $\mathcal{Q}(\mathbf{b}_i, I^*)$ assesses bounding box quality against established criteria, and $\mathcal{C}(\mathbf{b}_i, I)$ determines defect authenticity versus artifacts or normal structures.

The bounding box assessment component introduces precise geometric evaluation criteria through a multi-dimensional quality function:
\begin{equation}
\mathcal{Q}(\mathbf{b}_i, I^*) = \begin{cases}
\text{TIGHT} & \text{if } F_i \geq \tau_T \land G_i = 1 \\
\text{ACCEPTABLE} & \text{if } \tau_A \leq F_i < \tau_T \land G_i \geq 0.95 \\
\text{REFINEMENT} & \text{if } F_i < \tau_A \lor G_i < 0.95 \\
\text{MISALIGNED} & \text{if } H_i = 0
\end{cases}
\end{equation}
where $F_i = \text{Fit}(\mathbf{b}_i)$ measures spatial efficiency, $G_i = \text{Coverage}(\mathbf{b}_i)$ evaluates completeness, $H_i = \text{Alignment}(\mathbf{b}_i)$ assesses centering quality, and $\tau_T$ and $\tau_A$ are quality thresholds determined by the LMM agent based on visual assessment of bounding box fit quality. For cases requiring refinement, the system generates improved coordinates $\mathbf{b}_i^* = (x^*, y^*, w^*, h^*)$ through optimization:
\begin{equation}
\mathbf{b}_i^* = \arg\max_{\mathbf{b}} \left[ \beta \cdot F(\mathbf{b}) + \gamma \cdot G(\mathbf{b}) + \delta \cdot H(\mathbf{b}) \right]
\end{equation}
where $\beta$, $\gamma$, and $\delta$ are importance weights for spatial efficiency, coverage completeness, and alignment quality respectively, dynamically determined by the LMM agent.

The false positive elimination stage implements systematic rejection criteria based on established NDT knowledge and artifact identification protocols. The rejection function incorporates multiple validation checks:
\begin{equation}
R(d_i) = \mathcal{A}_i \lor S_i \lor N_i \lor E_i \lor P_i
\end{equation}
where each indicator $\mathcal{A}_i, S_i, N_i, E_i, P_i \in \{\text{True}, \text{False}\}$ represents LMM agent assessments for specific false positive categories: $\mathcal{A}_i$ identifies imaging artifacts and systematic patterns, $S_i$ recognizes normal structural features, $N_i$ detects random noise patterns, $E_i$ identifies boundary effects, and $P_i$ applies physical reasonableness constraints. The conservative evaluation principle is enforced through:
\begin{equation}
\text{Decision}(d_i) = \begin{cases}
\text{REJECT} & \text{if } R(d_i) = \text{True} \\
\text{UNCERTAIN} & \text{if } \epsilon_i < \tau_E \land R(d_i) = \text{False} \\
\text{CONFIRM} & \text{if } \epsilon_i \geq \tau_E \land R(d_i) = \text{False}
\end{cases}
\end{equation}
where $\epsilon_i = \text{Evidence}(d_i)$ represents evidence strength and $\tau_E$ represents the evidence confidence threshold established by the LMM agent based on visual analysis quality.

The confidence recalibration stage adjusts detector-provided confidence scores based on comprehensive visual evidence assessment and uncertainty quantification. The recalibration function incorporates evidence strength and consistency measures:
\begin{equation}
c_i^{\text{refined}} = c_i \cdot \omega_E(d_i) \cdot \omega_C(d_i) \cdot \omega_X(d_i)
\end{equation}
where $\omega_E(\cdot) \in [0,1]$, $\omega_C(\cdot) \in [0,1]$, and $\omega_X(\cdot) \in [0,1]$ are weighting functions computed by the LMM agent based on evidence strength, internal consistency, and contextual reasonableness assessments respectively.

The quality assurance stage implements multi-level verification through consistency checking, completeness evaluation, and overall reasonableness assessment. The comprehensive validation function ensures:
\begin{equation}
\text{QA}(\mathcal{D}) = \Phi_C(\mathcal{D}) \land \Phi_M(\mathcal{D}) \land \Phi_R(\mathcal{D})
\end{equation}
where $\Phi_C$, $\Phi_M$, and $\Phi_R$ represent consistency, completeness, and reasonableness validation components respectively.

The final output generation stage produces structured results incorporating refined detection coordinates, confidence scores, quality assessments, and comprehensive reasoning traces. The output function generates:
\begin{equation}
\text{Output}(\mathcal{D}) = \begin{cases}
\mathcal{D}^C & = \{d_i : \text{Decision}(d_i) = \text{CONFIRM}\} \\
\mathcal{D}^U & = \{d_i : \text{Decision}(d_i) = \text{UNCERTAIN}\} \\
\mathcal{D}^R & = \{d_i : \text{Decision}(d_i) = \text{REJECT}\} \\
\mathcal{M}^Q & = \{q_I, q_A, q_L\} \\
\mathcal{S} & = \{|\mathcal{D}^C|, |\mathcal{D}^U|, |\mathcal{D}^R|\}
\end{cases}
\end{equation}
where $\mathcal{D}^C$, $\mathcal{D}^U$, and $\mathcal{D}^R$ represent confirmed, uncertain, and rejected detection sets respectively, $\mathcal{M}^Q$ denotes the quality metrics set containing image quality ($q_I$), analysis confidence ($q_A$), and challenge assessments ($q_L$), and $\mathcal{S}$ represents the detection statistics set with counts of confirmed, uncertain, and rejected detections.

The EGR mechanism implements validation through prompt-engineered templates that guide the LMM agent through each analytical stage with specific attention to bounding box refinement capabilities, systematic false positive elimination, and comprehensive quality control measures. {These six stages are designed to function as a logically interdependent chain that emulates an expert's diagnostic workflow, and removing any single stage would disrupt the integrity of this cognitive process.} The framework establishes explicit reasoning traces for each detection while incorporating NDT domain expertise, enabling practitioners to understand the evidential basis for diagnostic conclusions and identify cases requiring additional expert review.

{To ensure the reproducibility of the EGR mechanism, particularly the dynamic determination of thresholds and weights, it is important to clarify that the LMM agent does not compute explicit numerical values for these parameters in a traditional algorithmic sense. Instead, these assessments are emergent properties of the LMM's reasoning process, guided by structured prompting. The LMM translates its qualitative visual analysis into semi-quantitative judgments. For instance, when determining the evidence confidence threshold $\tau_E$ or the recalibration weights $\omega_E$, $\omega_C$, $\omega_X$, the agent is prompted to evaluate visual evidence as ``strong,'' ``moderate,'' or ``weak'' based on features like defect clarity, contrast, and consistency with known patterns. These qualitative labels are then mapped to predefined decision pathways within the prompt logic. For example, a ``strong'' evidence rating directly leads to a CONFIRM decision, whereas a ``weak'' rating might trigger an UNCERTAIN classification. This approach replaces hard-coded numerical thresholds with a more flexible, context-aware reasoning framework. The mathematical notation in the EGR formulation serves to represent the reasoning process and decision flow in a structured form, rather than to specify predetermined threshold values.}

{For run-to-run consistency and deterministic outputs, the decoding temperature of the LMM is set to 0 during inference, which ensures that for the same input image and detection proposals, the EGR mechanism will consistently produce identical reasoning traces and final reports.}

Through this comprehensive evidence-grounded methodology, the system achieves enhanced diagnostic reliability with improved spatial accuracy, systematic false positive reduction, and robust quality assurance mechanisms. The structured validation process ensures that all detection outputs undergo thorough evidential examination with explicit consideration of geometric precision, physical reasonableness, and diagnostic confidence calibration, significantly improving overall system trustworthiness in automated X-ray defect detection applications.

\subsection{Training and Optimization}

The \textsc{InsightX Agent} employs a multi-stage training paradigm to meet the need for different parts of the agent. The training process initiates with visual perception optimization through supervised learning of the SDMSD architecture. The detector utilizes set-based loss functions incorporating Hungarian matching algorithms to establish an optimal assignment between predictions and ground truth annotations. The detection loss function integrates multiple complementary objectives:
\begin{equation}
\mathcal{L}_{\text{detection}} = \lambda_{\text{cls}} \mathcal{L}_{\text{cls}}(\hat{y}, y) + \lambda_{\text{bbox}} \mathcal{L}_{\text{bbox}}(\hat{\mathbf{b}}, \mathbf{b}) + \lambda_{\text{giou}} \mathcal{L}_{\text{giou}}(\hat{\mathbf{b}}, \mathbf{b})
\end{equation}
where $\hat{y}$ and $\hat{\mathbf{b}}$ denote predicted class labels and bounding boxes for refined sparse proposals, while $y$ and $\mathbf{b}$ represent ground truth annotations, and $\lambda_{\text{cls}}$, $\lambda_{\text{bbox}}$, and $\lambda_{\text{giou}}$ are loss weighting hyperparameters empirically set to balance different training objectives. Auxiliary losses are applied to initial dense proposals before NMS to enhance feature learning and representation quality.

Subsequently, the framework implements domain-specific knowledge injection through next-token prediction training on the NDT corpora. The language model component learns contextually appropriate response generation through the autoregressive objective:
\begin{equation}
\mathcal{L}_{\text{LM}} = -\sum_{i=1}^{N} \log P_\theta(t_i | t_{<i}, V)
\end{equation}
where $t_i$ represents the $i$-th token, $V$ denotes visual features from the perception module, and $\theta$ encompasses model parameters. This phase incorporates specialized domain terminology, standardized inspection protocols, and diagnostic reasoning patterns essential for NDT analysis, enabling the system to develop domain expertise.

The final training phase adapts the integrated system for structured output generation and reflection-based validation through instruction tuning methodologies. This process aligns model outputs with desired response formats while incorporating the reflection protocol. The refined training objective combines standard language modeling loss with format consistency regularization:
\begin{equation}
\mathcal{L}_{\text{fine-tune}} = \mathcal{L}_{\text{LM}} + \lambda_{\text{format}} \mathcal{L}_{\text{format}}(\hat{s}, s_{\text{template}})
\end{equation}
where $\mathcal{L}_{\text{format}}(\hat{s}, s_{\text{template}})$ enforces adherence to structured output templates between generated responses $\hat{s}$ and expected format templates $s_{\text{template}}$, and $\lambda_{\text{format}}$ is a weighting hyperparameter for format consistency.

The optimization process employs gradient accumulation and mixed-precision training strategies to manage the computational demands of large-scale multimodal architectures. Visual encoder and language model parameters are initialized from pre-trained weights and fine-tuned with LoRA to preserve learned representations.

Through this systematic training methodology, the \textsc{InsightX Agent} achieves integration of specialized visual perception capabilities with domain-aware language generation, resulting in a system capable of producing accurate, well-reasoned diagnostic outputs with enhanced reliability and interpretability for industrial NDT applications.

\section{Experiments and Discussions}
\subsection{Experimental Setup}

For experimental validation of the \textsc{InsightX Agent}, we utilized the casting inspection subset from the GDXray+ radiographic imaging collection \cite{mery_gdxray_2015}. This specialized dataset focuses on automated defect detection in industrial aluminum casting components, comprising automotive parts such as wheels and knuckles that were acquired through image intensifier technology.

{The dataset is organized into 67 distinct series, each corresponding to a specific casting part or inspection batch. To ensure a robust and practical evaluation, we first preprocessed the data by removing any series that were unlabeled or contained too few samples for a meaningful split. From the remaining series, we created our training and testing sets using an intra-series partitioning strategy. For each individual series, we randomly partitioned its images into an 80\% training and 20\% testing subset. The final training set (571 images) and testing set (143 images) were formed by aggregating the respective subsets from all series. This partitioning ensures that no data leakage occurs, as no single image appears in both sets, and the model is rigorously evaluated on its ability to generalize to unseen images within each part series it has been trained on, which closely mimics real-world industrial inspection scenarios. Each annotation includes precise bounding box coordinates for defect localization.}

For the following evaluations, we employ three key metrics common in object detection tasks to assess detection performance: precision, recall, and F1-score. Precision measures the accuracy of positive predictions, recall quantifies the ability to identify all actual defects and F1-score provides a harmonic mean of these two metrics. They are mathematically defined as:
\begin{equation}
\text{Precision} = \frac{TP}{TP + FP}
\end{equation}
\begin{equation}
\text{Recall} = \frac{TP}{TP + FN}
\end{equation}
\begin{equation}
\text{F1-score} = 2 \times \frac{\text{Precision} \times \text{Recall}}{\text{Precision} + \text{Recall}}
\end{equation}
where $TP$ represents true positives, $FP$ denotes false positives, and $FN$ represents false negatives. For all evaluations, we used an IoU threshold of 0.5 to determine whether a predicted bounding box matches a ground truth annotation.

{Moreover, to provide comprehensive analysis across different defect scales, we categorize detected objects into three size groups based on their bounding box area in pixels, following the standard COCO evaluation criteria: small (area $< 32^2$ pixels), medium ($32^2 \leq \text{area} < 96^2$ pixels), and large (area $\geq 96^2$ pixels) objects. These absolute pixel area thresholds are applied directly to the images at their evaluation resolution and do not depend on normalization relative to image size. This categorization is particularly useful for analyzing the results in casting defect detection, as different anomaly types exhibit distinct size characteristics.}

The implementation of \textsc{InsightX Agent} was developed using the PyTorch framework, with the HuggingFace Transformers library utilized for model architecture and training utilities. All experiments were conducted on a single NVIDIA A10 GPU with 24GB of VRAM. 

The foundation LMM used in the experiments was initialized with the weights of Qwen2.5-VL \cite{bai_qwen25-vl_2025}, which comprises 7 billion parameters. During the fine-tuning process, the LoRA rank was set to 8, with a scaling factor of 32. The learning rate was initialized at $1 \times 10^{-5}$, and a linear warm-up schedule was employed for the first 5\% of total training steps to ensure stability. The batch size was set to 32, with gradient accumulation over 16 steps to maximize GPU utilization while maintaining memory efficiency.

\subsection{Effectiveness of SDMSD and EGR}
\begin{table*}[htbp]
\centering
\caption{{Ablation Study Results to Validate the Effectiveness of SDMSD and EGR. Results are reported as Mean $\pm$ Std. Dev. (\%) over 3 independent runs.}}
\label{tab:ablation}
\resizebox{\linewidth}{!}{%
\begin{tabular}{@{}cccccccccc@{}}
\toprule
\multirow{2}{*}{Methods} & \multicolumn{4}{c}{Precision} & \multicolumn{4}{c}{Recall} & \multirow{2}{*}{F1-score} \\ 
\cmidrule(lr){2-5} \cmidrule(lr){6-9}
& Overall & Small & Medium & Large & Overall & Small & Medium & Large & \\ 
\midrule
LMM Only & $1.18 \pm 1.23$ & $0.08 \pm 0.06$ & $1.27 \pm 0.75$ & $3.08 \pm 2.11$ & $7.86 \pm 3.14$ & $1.47 \pm 0.98$ & $14.03 \pm 3.82$ & $9.77 \pm 2.54$ & $2.05 \pm 1.86$ \\
SDMSD Only & $92.68 \pm 1.07$ & $90.95 \pm 1.12$ & $93.36 \pm 0.91$ & $94.71 \pm 1.05$ & $98.94 \pm 0.73$ & $98.51 \pm 0.98$ & $98.54 \pm 0.65$ & $100.00 \pm 0.00$ & $95.71 \pm 0.66$ \\
InsightX Agent & $94.77 \pm 0.98$ & $91.49 \pm 0.73$ & $95.83 \pm 0.92$ & $97.80 \pm 1.21$ & $98.38 \pm 0.79$ & $97.08 \pm 0.67$ & $98.62 \pm 1.34$ & $100.00 \pm 0.00$ & $96.54 \pm 0.64$ \\ 
\bottomrule
\end{tabular}%
}
\end{table*}

To systematically evaluate the contribution of individual components within \textsc{InsightX Agent}, we conducted ablation studies across three configurations. The LMM Only configuration employs direct visual grounding, where the LMM is trained on the same dataset and receives X-ray images to generate bounding box coordinates without specialized detection components. The SDMSD Only configuration utilizes the Sparse Deformable Multi-Scale Detector as a standalone system without EGR validation. The complete \textsc{InsightX Agent} framework integrates the LMM agent core with SDMSD tool invocation followed by EGR-based validation.

{Results in \autoref{tab:ablation} reveal significant performance variations across configurations. The LMM Only approach achieves $1.18 \pm 1.23\%$ overall precision and $7.86 \pm 3.14\%$ recall, with particularly poor small defect performance ($0.08 \pm 0.06\%$ precision, $1.47 \pm 0.98\%$ recall). The high standard deviations reflect the inherent instability of direct multimodal grounding approaches for precise spatial localization tasks. This demonstrates the limitations of direct multimodal grounding for precise spatial localization in industrial imagery, validating the necessity of specialized detection architectures.}

{The SDMSD Only configuration exhibits superior performance with $92.68 \pm 1.07\%$ overall precision and $98.94 \pm 0.73\%$ recall, demonstrating the effectiveness of sparse deformable attention mechanisms and multi-scale feature processing. This validates the architectural design choices, particularly the dense-to-sparse proposal generation strategy optimized for small and densely distributed anomalies in radiographic imagery.}

{The complete \textsc{InsightX Agent} framework demonstrates precision improvement from $92.68\%$ to $94.77\%$ ($2.09\%$ increase) with modest recall reduction from $98.94\%$ to $98.38\%$ ($0.56\%$ decrease). This precision-recall trade-off is pronounced for small defects, where precision improves from $90.95\%$ to $91.49\%$ while recall decreases from $98.51\%$ to $97.08\%$. The F1-score improvement from $95.71\%$ to $96.54\%$ indicates that precision gains outweigh recall reduction, enhancing overall detection quality. Notably, InsightX Agent exhibits the lowest standard deviations across all metrics ($\pm 0.64\%$ F1-score), indicating superior stability and robustness compared to both LMM Only and SDMSD Only configurations.}

{This performance pattern reflects the conservative validation behavior of the EGR mechanism. EGR effectively eliminates false positives through systematic visual verification and domain knowledge application, but occasionally rejects legitimate detections with subtle visual characteristics or ambiguous presentations. Such cases may include low-contrast defects, anomalies partially obscured by overlapping structures, or subtle dimensional irregularities challenging visual confirmation. The study validates the synergistic effectiveness of SDMSD and EGR components. SDMSD provides robust detection capabilities while EGR contributes essential validation functions that significantly enhance precision without substantially compromising sensitivity. This component interaction achieves reliable, interpretable NDT analysis with enhanced trustworthiness for industrial inspection applications.}

\subsection{Comparison with Existing Methods}
\begin{table*}[htbp]
\centering
\caption{{Performance Comparison of \textsc{InsightX Agent} with Existing Object Detection Methods on the GDXray+ Dataset. Results are reported as Mean $\pm$ Std. Dev. (\%) over 3 independent runs.}}
\label{tab:comparison}
\resizebox{\linewidth}{!}{%
\begin{tabular}{@{}cccccccccc@{}}
\toprule
\multirow{2}{*}{Methods} & \multicolumn{4}{c}{Precision} & \multicolumn{4}{c}{Recall} & \multirow{2}{*}{F1-score} \\ 
\cmidrule(lr){2-5} \cmidrule(lr){6-9}
& Overall & Small & Medium & Large & Overall & Small & Medium & Large & \\ 
\midrule
Faster R-CNN & $88.19 \pm 1.76$ & $82.94 \pm 2.43$ & $87.71 \pm 1.85$ & $95.78 \pm 1.42$ & $91.33 \pm 1.31$ & $88.49 \pm 1.62$ & $90.18 \pm 1.15$ & $94.68 \pm 1.03$ & $89.73 \pm 1.11$ \\
YOLOX-s & $94.61 \pm 1.03$ & $89.62 \pm 1.48$ & $96.68 \pm 1.15$ & $98.38 \pm 0.76$ & $96.93 \pm 0.85$ & $93.45 \pm 1.21$ & $98.54 \pm 0.62$ & $100.00 \pm 0.00$ & $95.76 \pm 0.67$ \\
DINO & $90.31 \pm 1.25$ & $87.14 \pm 1.53$ & $90.25 \pm 1.38$ & $95.06 \pm 1.18$ & $98.91 \pm 0.65$ & $98.50 \pm 0.77$ & $98.57 \pm 0.54$ & $100.00 \pm 0.00$ & $94.41 \pm 0.74$ \\
Deformable DETR & $81.77 \pm 2.13$ & $75.52 \pm 2.89$ & $83.24 \pm 2.41$ & $88.99 \pm 2.05$ & $92.05 \pm 1.72$ & $89.18 \pm 2.23$ & $91.57 \pm 1.63$ & $96.64 \pm 1.25$ & $86.61 \pm 1.42$ \\
PVTv2 & $89.78 \pm 1.33$ & $85.23 \pm 1.75$ & $90.61 \pm 1.46$ & $95.51 \pm 1.29$ & $96.37 \pm 0.95$ & $92.03 \pm 1.36$ & $98.52 \pm 0.71$ & $100.00 \pm 0.00$ & $92.96 \pm 0.84$ \\
InsightX Agent & $94.77 \pm 0.98$ & $91.49 \pm 0.73$ & $95.83 \pm 0.92$ & $97.80 \pm 1.21$ & $98.38 \pm 0.79$ & $97.08 \pm 0.67$ & $98.62 \pm 1.34$ & $100.00 \pm 0.00$ & $96.54 \pm 0.64$ \\ 
\bottomrule
\end{tabular}%
}
\end{table*}

To evaluate the performance of \textsc{InsightX Agent} against established object detection methods, we conducted comparative analysis with five representative approaches spanning different architectures. The baseline methods include Faster R-CNN \cite{ren_faster_2016} with ResNet-50 backbone and Feature Pyramid Network (FPN), representing two-stage detection architectures; YOLOX-s \cite{ge_yolox_2021}, exemplifying modern single-stage detectors with advanced training strategies; DINO \cite{caron_emerging_2021} with ResNet-50 and 4-scale processing, demonstrating transformer-based detection with query-based mechanisms; Deformable DETR \cite{zhu_deformable_2020} with ResNet-50, showcasing deformable attention in detection transformers; and PVTv2 \cite{wang_pvt_2022} integrated with RetinaNet, illustrating pyramid vision transformer architectures. {All baseline methods were trained and evaluated under identical experimental conditions using the same GDXray+ dataset partitions to ensure a fair comparison. To maintain consistency, all models, including our proposed SDMSD, utilized backbones pre-trained on ImageNet-1K. For training, we employed the AdamW optimizer with a cosine annealing learning rate scheduler and a batch size of 16. The initial learning rate was set to $2 \times 10^{-4}$ for all transformer-based models (DINO, Deformable DETR, and our SDMSD) and $1 \times 10^{-3}$ for other architectures, reflecting common practices. All models were trained for at most 100 epochs, which we found sufficient for convergence on this dataset. No data augmentation was applied during training.}

{To enhance statistical robustness and assess the reliability of performance differences, we conducted all experiments with three independent runs using different random seeds for model training and data shuffling. The reported results in all tables represent the mean performance across these runs, with standard deviations indicating the variability. This experimental protocol allows us to evaluate not only the average performance but also the consistency and stability of each method.}

{The comparative results presented in \autoref{tab:comparison} demonstrate the superior performance of \textsc{InsightX Agent} across multiple evaluation metrics. The proposed framework achieves the highest F1-score of $96.54 \pm 0.64\%$, representing improvements of $6.81\%$ over Faster R-CNN, $0.78\%$ over YOLOX-s, $2.13\%$ over DINO, $9.93\%$ over Deformable DETR, and $3.58\%$ over PVTv2. More significantly, \textsc{InsightX Agent} exhibits exceptional precision of $94.77 \pm 0.98\%$ while maintaining high recall of $98.38 \pm 0.79\%$, indicating an optimal balance between false positive elimination and comprehensive defect detection. The relatively low standard deviations across all metrics ($\pm 0.64\%$ F1-score, $\pm 0.98\%$ precision, $\pm 0.79\%$ recall) demonstrate the superior stability and reliability of the agentic framework compared to conventional detection methods.}

{The data reveals distinctive patterns that illuminate the effectiveness of the agentic framework approach. Traditional methods such as Faster R-CNN and Deformable DETR demonstrate moderate precision but achieve reasonable recall, reflecting their design focus on comprehensive detection coverage. However, these approaches exhibit substantial performance degradation for small defects, with Faster R-CNN achieving only $82.94 \pm 2.43\%$ precision and Deformable DETR reaching merely $75.52 \pm 2.89\%$ precision for small-scale anomalies. The high standard deviations in these methods further indicate instability in handling challenging detection scenarios. This limitation stems from the inherent challenges of detecting subtle defects in radiographic imagery using conventional feature extraction and proposal generation mechanisms.}

{In contrast, modern single-stage detectors like YOLOX-s achieve competitive performance with $94.61 \pm 1.03\%$ precision and $96.93 \pm 0.85\%$ recall, demonstrating the effectiveness of advanced training strategies and architectural improvements. However, the performance gap between YOLOX-s and \textsc{InsightX Agent} reveals the fundamental advantage of the agentic approach: while YOLOX-s relies purely on learned feature representations, \textsc{InsightX Agent} incorporates systematic validation through EGR, enabling more reliable discrimination between genuine defects and imaging artifacts.}

{The transformer-based approaches, particularly DINO, exhibit interesting characteristics with exceptionally high recall ($98.91 \pm 0.65\%$) but moderate precision ($90.31 \pm 1.25\%$). This pattern suggests that query-based detection mechanisms effectively identify potential defect regions but may generate false positives due to the challenging nature of radiographic imagery interpretation. The superior precision of \textsc{InsightX Agent} demonstrates the critical value of the EGR mechanism in systematic false positive elimination through structured visual verification and domain knowledge application.}

Beyond quantitative performance metrics, \textsc{InsightX Agent} provides unique advantages that distinguish it from conventional detection approaches. The agentic framework enables interpretable diagnostic reasoning through structured reflection protocols, allowing operators to understand the evidential basis for each detection decision. This interpretability is particularly crucial in industrial NDT applications where diagnostic transparency directly impacts operational safety and regulatory compliance. Furthermore, the framework's interactive capabilities enable dynamic query processing and contextual analysis, transforming static detection outputs into comprehensive diagnostic dialogues that support informed decision-making across varying operator expertise levels. The qualitative analysis presented in the following section will further demonstrate these advantages through case studies and visual evidence.

\subsection{Qualitative Analysis}

\subsubsection{Comparison with Conventional Approaches}

\begin{figure*}[htbp]
\centering
\includegraphics[width=\linewidth]{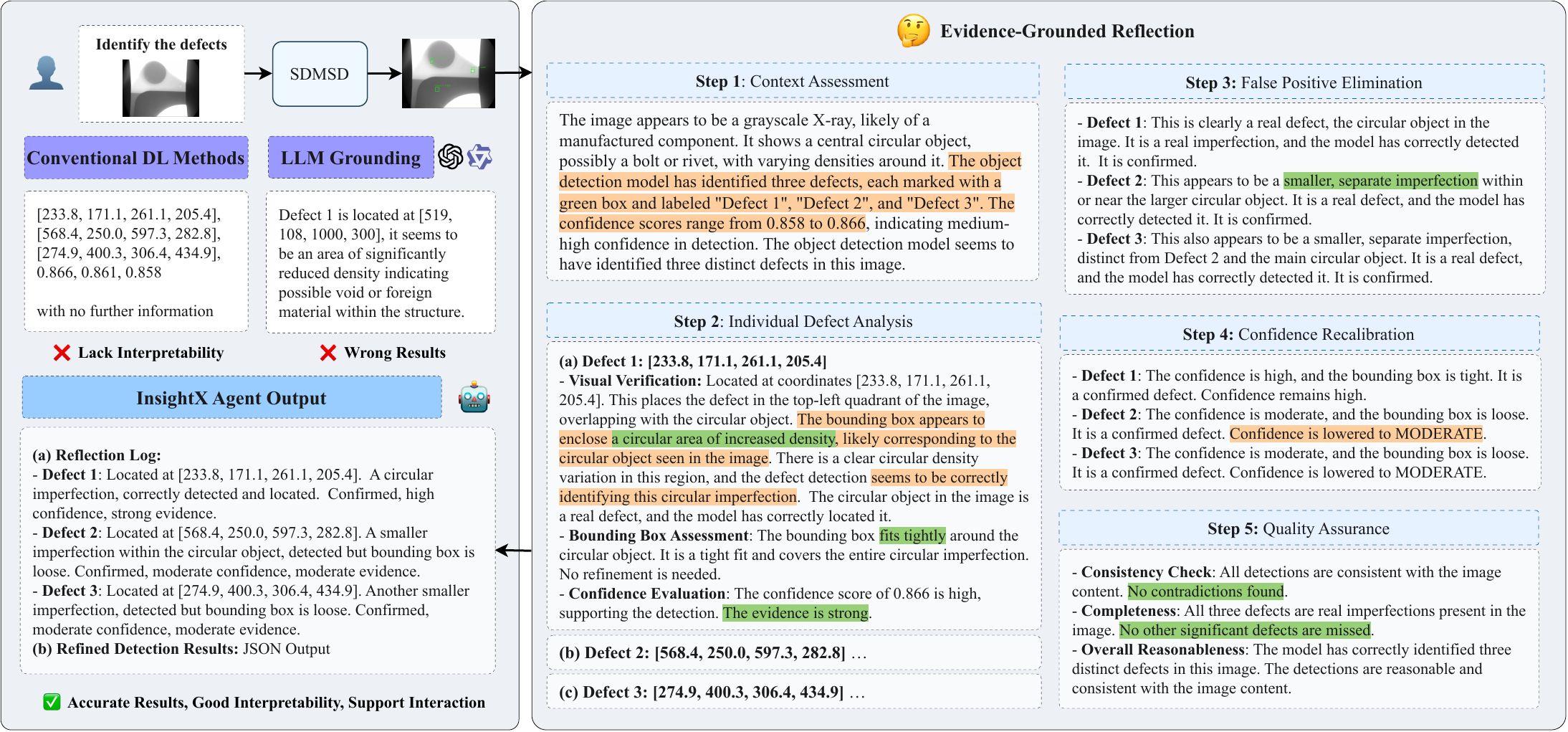} 
\caption{Comparative analysis of NDT diagnostic approaches demonstrating the transformative capabilities of \textsc{InsightX Agent}.}
\label{fig:qualitative_comparison}
\end{figure*}

To comprehensively evaluate the transformative impact of \textsc{InsightX Agent} beyond quantitative metrics, we conducted a qualitative analysis examining the interpretability, reasoning transparency, and practical applicability of the proposed agentic approach. \autoref{fig:qualitative_comparison} presents a representative case study that illuminates the fundamental paradigmatic differences between conventional detection methods, direct LMM grounding approaches, and the proposed agentic framework.

Conventional deep learning methods, exemplified by standard object detectors, provide accurate spatial localization through precise bounding box coordinates and confidence scores. However, these approaches generate purely numerical outputs such as [233.8, 171.1, 261.1, 205.4] with corresponding confidence values, which would require substantial expert interpretation and validation to derive actionable insights. The lack of explanatory context severely limits operator understanding of the detection rationale, particularly for personnel with varying levels of NDT expertise. This interpretability deficit represents a critical barrier to trust establishment and informed decision-making in industrial inspection workflows.

Direct LMM grounding approaches attempt to address interpretability limitations by leveraging LMMs for simultaneous detection and explanation generation. While these methods provide contextual descriptions such as ``Defect is located at [519, 108, 1000, 300], it seems to be an area of significantly reduced density indicating possible void or foreign material within the structure,'' they suffer from substantial localization inaccuracies and hallucination artifacts. The spatial coordinates frequently exhibit significant deviations from ground truth positions, undermining the reliability essential for critical industrial applications. Furthermore, the generated explanations, while linguistically coherent, may lack evidential grounding in actual visual features, leading to plausible but factually incorrect diagnostic assessments.

In contrast, \textsc{InsightX Agent} demonstrates a fundamentally transformed diagnostic paradigm through its call of tools and use of EGR. The system generates comprehensive analytical reports after five systematic validation stages: context assessment, individual defect analysis, false positive elimination, confidence recalibration, and quality assurance.

The context assessment stage establishes fundamental understanding through systematic characterization of the input image and identification of key structural features, providing essential background information that informs subsequent analytical processes. The individual defect analysis stage implements detailed evidential reasoning, where each detected defect undergoes comprehensive evaluation encompassing visual verification, spatial localization assessment, and confidence quantification to ensure diagnostic conclusions are explicitly grounded in observable visual evidence. The false positive elimination stage implements systematic verification protocols through assessment, in which the system confirms genuine defects while maintaining conservative evaluation standards that prioritize diagnostic reliability over detection sensitivity. Finally, confidence recalibration and quality assurance stages implement sophisticated self-assessment mechanisms, where the system dynamically adjusts confidence scores based on visual evidence strength and conducts comprehensive consistency verification to ensure diagnostic reliability while providing operators with calibrated confidence estimates that reflect actual evidence quality. Following this systematic validation process, the framework generates structured outputs comprising reflection logs and JSON-formatted defect results.

Beyond structured reasoning, \textsc{InsightX Agent} enables interactive diagnostic capabilities that transform static detection outputs into dynamic analytical dialogues. Operators can query the system for clarification on ambiguous findings, request additional analysis of specific regions, or seek contextual information regarding defect implications. This interactivity addresses the diverse expertise levels encountered in industrial settings, enabling both novice operators to receive detailed explanatory guidance and experienced inspectors to efficiently access targeted analytical insights.

{The qualitative analysis demonstrates that \textsc{InsightX Agent} represents a fundamental paradigmatic shift from passive detection systems to active diagnostic reasoning frameworks. The integration of specialized detection capabilities with evidence-grounded validation mechanisms creates a synergistic system that achieves superior accuracy while offering significantly improved interpretability and trustworthiness compared to conventional ``black box'' detection approaches. This transformation addresses critical limitations in current NDT automation approaches, establishing a foundation for enhanced operator confidence and more informed decision-making in safety-critical industrial applications.}

{
\begin{figure*}[htbp]
\centering
\includegraphics[width=\linewidth]{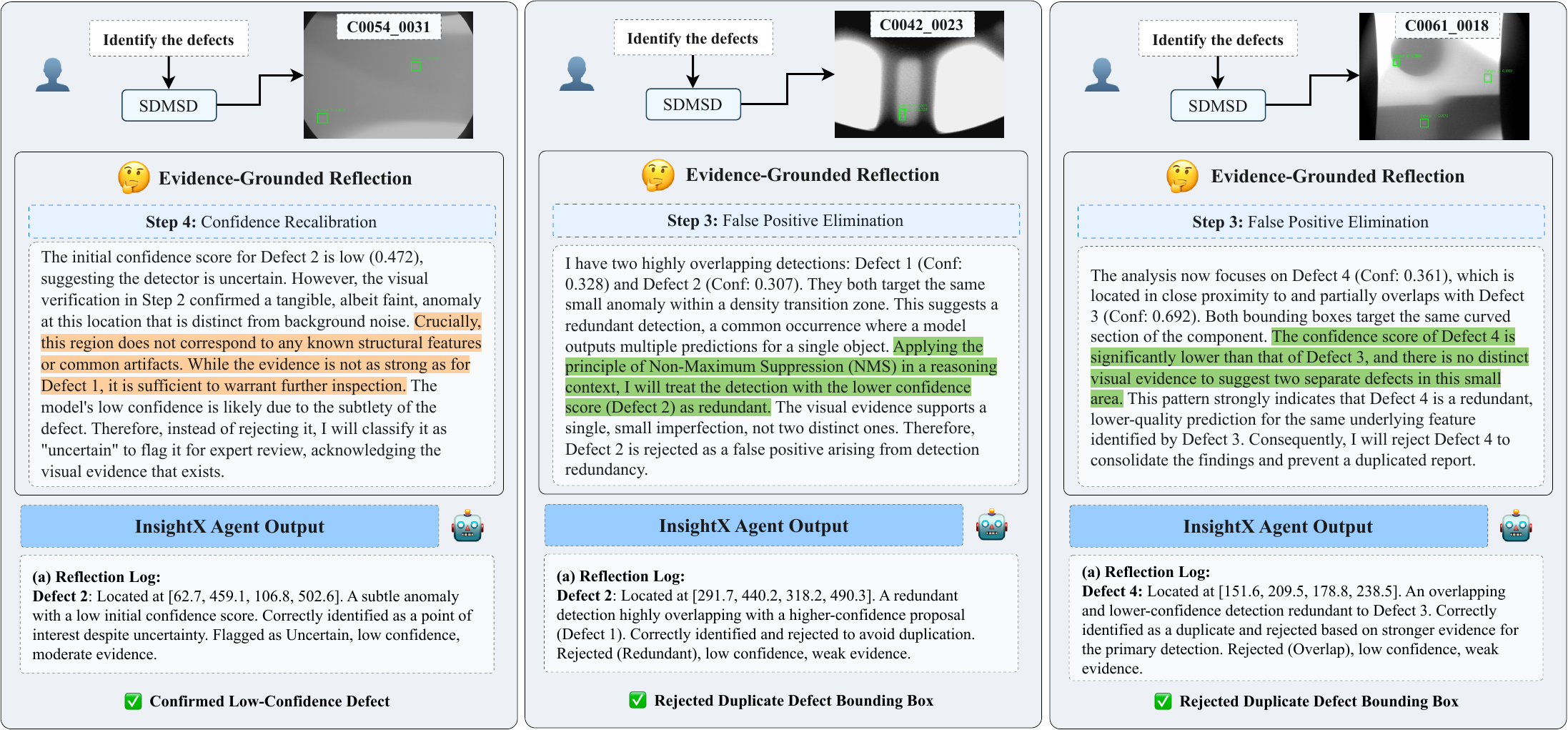}
\caption{Case studies of the Evidence-Grounded Reflection (EGR) mechanism handling complex and ambiguous scenarios.}
\label{fig:egr_cases}
\end{figure*}
}

\subsubsection{EGR Case Study}

{To further illustrate the superiority of EGR and our agentic approach, \autoref{fig:egr_cases} presents three case studies demonstrating how this mechanism navigates complex and ambiguous scenarios that often challenge conventional detectors.}

{In the first case (left), the SDMSD identifies a subtle anomaly with a low initial confidence score (0.472). A standard pipeline might discard this detection based on a simple confidence threshold. However, the \textsc{InsightX Agent} proceeds to a deeper analysis. During EGR, it verifies that the faint visual evidence does not correspond to any known structural features or artifacts. Acknowledging the subtlety but also the validity of the evidence, the agent intelligently recalibrates its assessment, classifying the defect as ``Uncertain'' and flagging it for expert review. This prevents a potential false negative by handling uncertainty in a safe and transparent manner.}

{The second and third cases (center and right) showcase the agent's ability to resolve detection redundancy, a common issue where a single defect triggers multiple bounding box proposals. In the center case, the agent identifies two highly overlapping boxes targeting the same small anomaly. Instead of blindly reporting both, it correctly identifies the lower-confidence detection as a redundant result and rejects it. Similarly, in the right case, the agent analyzes two partially overlapping detections. It reasons that the confidence of one is significantly lower and there is no distinct visual evidence to support the presence of two separate defects. It then correctly rejects the weaker proposal to consolidate the finding. These examples demonstrate that \textsc{InsightX Agent} moves beyond the rigid, purely algorithmic post-processing of conventional detectors, instead employing a flexible, evidence-based reasoning process to enhance the reliability and clarity of its final diagnostic output.}

\subsection{Computational Cost and Scalability}

{In the context of industrial applications, understanding the computational cost and inference latency of \textsc{InsightX Agent} is crucial for evaluating its practical deployment feasibility. Our framework's inference process consists of two main stages: defect detection by the SDMSD and evidence-grounded reflection by the LMM agent.}

{The SDMSD, being a highly optimized convolutional and transformer-based detector, is computationally efficient. On an NVIDIA A10 GPU, it processes a single X-ray image within 1 second. This stage is highly parallelizable and can benefit significantly from batching in a production environment, where multiple images can be processed simultaneously to increase throughput.}

{The EGR stage, however, introduces a notable latency overhead, as it involves an autoregressive token generation process by the 7B-parameter LMM. The time required for EGR depends on the number of initial detections and the complexity of the reasoning required. For a typical image with 3-5 defect proposals, the EGR process takes approximately 45-60 seconds to generate the full reflection log and structured output. Consequently, the total inference time per image for the full \textsc{InsightX Agent} pipeline could range depending on the depth and detail of the reflection process.}

{This demonstrates the trade-off between real-time performance and reliability. While the \textsc{InsightX Agent} may not be suitable for applications requiring real-time inspection at high frame rates, this latency performance is sufficient for many critical NDT scenarios, such as batch-based quality control or in-depth analysis of high-value components, where diagnostic accuracy and transparency are prioritized over raw processing speed. For industrial deployment scenarios requiring higher throughput, multi-GPU parallel processing can be employed to analyze multiple images concurrently. Furthermore, optimization techniques such as knowledge distillation to smaller LMM architectures and model quantization could substantially reduce the EGR latency while maintaining diagnostic quality, as discussed in our Future Work section.}

\section{Conclusions and Future Work}

This paper presents \textsc{InsightX Agent}, a novel LMM-based agentic framework that addresses critical limitations in automated X-ray NDT analysis. The proposed approach fundamentally transforms the paradigm from passive detection systems to active diagnostic reasoning frameworks, addressing key challenges in interpretability, reliability, and operator trust that have hindered the widespread adoption of AI-driven industrial inspection systems.

The core architectural innovation lies in positioning an LMM as a central orchestrator that intelligently utilizes a suite of integrated tools, including the Sparse Deformable Multi-Scale Detector (SDMSD) for precise anomaly localization and the Evidence-Grounded Reflection (EGR) mechanism for systematic validation. The SDMSD effectively addresses the challenging detection scenarios characteristic of X-ray imagery through its dense-to-sparse proposal generation strategy and deformable attention mechanisms, while the EGR mechanism enables autonomous validation, refinement, and uncertainty flagging through structured evidential scrutiny.

Experimental validation on the GDXray+ dataset demonstrates the superior performance of \textsc{InsightX Agent}, achieving a 96.54\% F1-score, outperforming state-of-the-art detection methods. {More importantly, through its evidence-grounded reflective reasoning mechanisms, the framework offers significantly improved interpretability and trustworthiness compared to traditional ``black box'' detection approaches.} The ablation studies confirm the synergistic effectiveness of the integrated components, with the EGR mechanism contributing a 2.09\% precision improvement through systematic false positive elimination, albeit with conservative evaluation characteristics that occasionally reject subtle but genuine defects.

The qualitative analysis reveals the transformative impact of the agentic approach, demonstrating how the framework transcends conventional detection outputs to provide comprehensive diagnostic reasoning with explicit evidential grounding. Unlike traditional methods that generate purely numerical coordinates requiring expert interpretation, or direct LMM approaches that suffer from localization inaccuracies and hallucination artifacts, \textsc{InsightX Agent} is able to deliver structured analytical reports that integrate visual verification, domain knowledge, and systematic validation.

Furthermore, the significance of this work extends beyond performance improvements to a fundamental paradigmatic transformation in industrial NDT automation. By enabling interactive diagnostic capabilities, the framework addresses diverse operator expertise levels and transforms static detection outputs into dynamic analytical dialogues. This advancement is particularly crucial for safety-critical applications where diagnostic transparency directly impacts operational safety and regulatory compliance.

{Despite its promising performance, the proposed framework also demonstrates limitations in certain aspects. Firstly, the agentic reasoning process, particularly the iterative EGR mechanism, introduces computational latency. Unlike feed-forward detectors that provide near-instantaneous outputs, \textsc{InsightX Agent} engages in a token-by-token reflective dialogue, which prioritizes diagnostic reliability and interpretability over raw processing speed. This trade-off may present challenges for deployment in high-throughput, real-time inspection pipelines where latency is a critical constraint. Secondly, the current model's effectiveness has been validated specifically on aluminum casting datasets. Its generalization capabilities to other materials, such as steel welds or composite structures, or different inspection scenarios like electronics manufacturing, remain unevaluated. The visual characteristics of defects and the underlying domain knowledge can vary significantly across these contexts, which would necessitate further domain-specific fine-tuning for both the SDMSD detector and the LMM agent core. {Finally, while we demonstrate the framework's enhanced interpretability through reflection logs and case studies, the current study lacks a formal user evaluation. The practical benefits, such as reduced operator validation time or increased diagnostic confidence, have not been quantitatively measured through human-in-the-loop experiments.}}

{Future work can be directed toward addressing these limitations and expanding the agent's capabilities. To mitigate computational latency, we could explore model optimization techniques, including knowledge distillation to a smaller, faster agent, quantization, and speculative decoding for the LMM. Furthermore, few-shot and zero-shot learning approaches can be employed to enable the agent to rapidly adapt to new, unseen inspection tasks with minimal additional data. {Another key priority for future work will be to conduct a formal user study. This will be crucial for assessing the real-world utility and user acceptance of this new approach.}}

\printbibliography

\vfill

\end{document}